\documentclass{article}

\usepackage{microtype}
\usepackage{graphicx}
\usepackage{subfigure}
\usepackage{amssymb}
\usepackage{amsmath}
\usepackage{amsthm}
\usepackage{booktabs} 
\usepackage{tabularx}

\usepackage{tabularx,ragged2e}

\newcommand{\path}{\mathbf{l}}

\usepackage{hyperref}

\usepackage[accepted]{icml2019}

\begin{document}

\twocolumn[
\icmltitle{PAN: Path Integral Based Convolution for Deep Graph Neural Networks}



\icmlsetsymbol{equal}{*}

\begin{icmlauthorlist}
\icmlauthor{Zheng Ma}{pu}
\icmlauthor{Ming Li}{scnu,ltu}
\icmlauthor{Yu Guang Wang}{unsw}

\end{icmlauthorlist}

\icmlaffiliation{ltu}{Department of Computer Science and Information Technology, La Trobe University, Melbourne, Australia}
\icmlaffiliation{scnu}{School of Information Technology in Education, South China Normal University, Guangzhou, China}
\icmlaffiliation{unsw}{School of Mathematics and Statistics, The University of New South Wales, Sydney, Australia}
\icmlaffiliation{pu}{Department of Physics, Princeton University, New Jersey, USA}

\icmlcorrespondingauthor{Zheng Ma}{zhengm@princeton.edu}

\icmlkeywords{Graph Neural Networks, Maximal entropy, Graph Convolution, Geometric Deep Learning}

\vskip 0.3in
]



\printAffiliationsAndNotice{}  

\begin{abstract}
Convolution operations designed for graph-structured data usually utilize the graph Laplacian, which can be seen as message passing between the adjacent neighbors through a generic random walk. In this paper, we propose PAN, a new graph convolution framework that involves every path linking the message sender and receiver with learnable weights depending on the path length, which corresponds to the maximal entropy random walk. PAN generalizes the graph Laplacian to a new transition matrix we call \emph{maximal entropy transition} (MET) matrix derived from a path integral formalism. Most previous graph convolutional network architectures can be adapted to our framework, and many variations and derivatives based on the path integral idea can be developed. Experimental results show that the path integral based graph neural networks have great learnability and fast convergence rate, and achieve state-of-the-art performance on benchmark tasks.
\end{abstract}

\section{Introduction}
\label{intro}
The triumph of convolutional neural networks (CNNs) has motivated researchers to develop similar architectures for graph-structured data. The problem is challenging due to the absence of regular grids. One notable proposal is to define convolutions in the Fourier space \cite{BrZaSzLe2013,Bronstein_etal2017}. This method relies on finding the spectrum of the graph Laplacian $I-D^{-1}A$ or $I-D^{-\frac{1}{2}}AD^{-\frac{1}{2}}$ (depending on how normalization is done), where $A$ is the adjacency matrix of the graph and $D$ is the corresponding degree matrix, and then applies filters to the components of input signal $X$ under the basis of the graph Laplacian. Due to the high computational complexity of diagonalizing the graph Laplacian, many simplifications have been proposed \cite{ChebNet,gcn}.

The graph Laplacian based methods essentially rely on message passing \cite{gilmer2017neural} between directly connected nodes with equal weights shared among all edges, which is at the heart a generic random walk (GRW) defined on graphs. This can be seen most obviously from the GCN model \cite{gcn}, where the normalized adjacency matrix is directly applied to the left hand side of the input. Mathematically, $D^{-1}A$ is known as the transition matrix of a particle doing a random walk on the graph, where the particle hops to all directly connected nodes with equiprobability. Many direct space based methods \cite{LiTaBrZe2015,node2vec_2016,Planetoid_2016,gat,Attention_GCN} can be viewed as generalizations of GRW that enable one to do a biased average of the neighbors, although they are in general more complicated and the bias usually depends on trainable parameters.

In this paper, we present PAN, a general framework for graph convolution inspired by the path integral idea in physics. We go beyond the generic diffusion picture and consider the message passing along all possible paths between the sender and receiver on a graph, with trainable weights depending on the path length. This results in a
\emph{maximal entropy transition} (MET) matrix, which plays the same role as graph Laplacian. By introducing a fictitious temperature, we can continuously tune our model from a fully localized one (i.e., MLP) to a global structure based model. Great learnability and fast convergence rate of PAN are observed when training the benchmark dataset. Numerous variations of MET can be developed, and many current models can be seen as special cases of the presented framework.

\section{Model}
In the most general form, we heuristically propose a statistical mechanics model on how information is averaged between different nodes on a given graph. Using the formalism of Feynman's path integral \cite{feynman1979path}, but modified for discrete graph structures, we write observable $\phi_i$ at node $i$ for a graph with $N$ nodes as
\begin{equation} \label{eq:stat}
\phi_i=\frac{1}{Z_i}\sum_{j=1}^{N}\phi_j \sum_{\{\mathbf l|l_0=i,l_{|\path|}=j\}}e^{-\frac{E[\mathbf l]}{T}},
\end{equation}
where $Z_i$ is the normalization factor known as the \textit{partition function} for the $i$-th node, and $\mathbf l$ is any path formed on the graph. Here a path $\path$ is a sequence of connected nodes $(l_0l_1\dots l_{|\path|})$ where $A_{l_il_{i+1}}=1$, and the length of the path is denoted by $|\path|$. Since a statistical mechanics perspective is more straightforward in our case, we directly change the exponential term, which is originally an integral of Lagrangian, to a Boltzmann's factor with fictitious energy $E[\path]$ and temperature $T$. (We choose Boltzmann's constant $k_B=1$.) Nevertheless, we still exploit the fact that the energy is a functional of the path, which gives us a way to weight the influence of other nodes through a certain path. The fictitious temperature controls the excitation level of the system, which reflects that to what extent information is localized or extended. Specifically, low temperature corresponds to a low-pass filter, while high temperature corresponds to a high-pass one. In practice, there is no need to learn the fictitious temperature or energy separately, instead the neural networks can directly learn the overall weights, as would be made clearer later.

To obtain an explicit form of our model, we now introduce some mild assumptions and simplifications. Intuitively, we know that information quality usually decays as the path between the message sender and the receiver becomes longer, thus it is reasonable to assume that the energy is not only a functional of path, but can be further simplified as a function that solely depends on the length of the path. In the random walk picture, this means that the hopping is equiprobable among all the paths that have the same length, which maximizes the Shannon entropy of the probability distribution of paths globally, and thus the random walk is given the name maximal entropy random walk \cite{burda2009localization}. For a weighted graph, a feasible choice for the functional form of the energy could be $E(l_{\rm eff})$, where the effective length of the path $l_{\rm eff}$ can be defined as a summation of the inverse of weights along the path, i.e., $l_{\rm eff}=\sum_{i=0}^{|l|-1}1/w_{l_il_{i+1}}$. After simplification, we can regroup the summation by first conditioning on the length of the path. Define the overall $n$-th layer weight $k(n;i)$ for node $i$ by
\begin{equation} \label{eq:kn}
k(n;i)=\frac{1}{Z_i}{\sum_{j=1}^{N}g(i,j;n)}e^{-\frac{E(n)}{T}},
\end{equation}
where $g(i,j;n)$ denotes the number of paths between nodes $i$ and $j$ with length of $n$, or \textit{density of states} for the energy level $E(n)$ with respect to nodes $i$ and $j$, and the summation is taken over all nodes of the graph. Presumably, the energy $E(n)$ is an increasing function of $n$, which leads to a decaying weight as $n$ increases.\footnote{This does not mean that $k(n;i)$ should necessarily be a decreasing function, since $g(i,j;n)$ grows exponentially in general. It would be valid to apply a cutoff as long as $E(n)\gg nT\ln \lambda_1$ for large $n$, where $\lambda_1$ is the largest eigenvalue of the adjacency matrix $A$.} By applying a cutoff of the maximal path length $L$, we can rewrite \eqref{eq:stat} as
\begin{align}
\phi_i&=\sum_{n=0}^{L}k(n;i)\sum_{j=1}^{N}\frac{g(i,j;n)}{\sum_{s=1}^{N}g(i,s;n)}\phi_j\notag\\
&=\frac{1}{Z_i}\sum_{n=0}^{L}e^{-\frac{E(n)}{T}}\sum_{j=1}^{N}g(i,j;n)\phi_j,
\label{eq:sumkn}
\end{align}
and the partition function can be explicitly written as
\begin{equation} \label{eq:partition}
Z_i=\sum_{n=0}^{L}e^{-\frac{E(n)}{T}}\sum_{j=1}^{N}g(i,j;n).
\end{equation}
A nice feature of this formalism is that we can easily compute $g(i,j;n)$ by raising the power of the adjacency matrix $A$ to $n$, which is a well-known property of the adjacency matrix from graph theory, i.e.,
\begin{equation} \label{eq:An}
g(i,j;n)=A^n_{ij}.
\end{equation}
Clearly, from \eqref{eq:sumkn} and \eqref{eq:An} we have a group of self-consistent equations governed by a transition matrix $M$ (a counterpart of the \textit{propagator} in quantum mechanics), which is defined as
\begin{equation} \label{eq:Propagator}
M_{ij}=\sum_{n=0}^{L}k(n;i)\frac{A^n_{ij}}{\sum_{s=1}^NA^n_{is}}.
\end{equation}
We call the matrix $M$ \emph{maximal entropy transition} (MET) matrix, with regard to the fact that it realizes maximal entropy under the microcanonical ensemble. This transition matrix replaces the role of the graph Laplacian under our formalism. It can be written in a more compact form
\begin{equation} \label{eq:Propagator2}
M=Z^{-1}\sum_{n=0}^{L}e^{-\frac{E(n)}{T}}A^n,
\end{equation}
where $Z={\rm diag}(Z_i)$. More generally, for paths with constraints such as shortest paths or self-avoiding paths, $A^n$ can be replaced by another matrix $G(n)$, where $G_{ij}(n)=g(i,j;n)$.

The \emph{eigenstates}, or the basis of the system $\{\psi_i\}$ satisfy
\begin{equation} \label{eq:eigen}
M\psi_i=\lambda_i\psi_i.
\end{equation}
Similar to the basis formed by the graph Laplacian, one can define graph convolution based on the new basis we obtained in \eqref{eq:eigen}, which has a distinct new physical meaning. The convolution associated with MET is computationally nontrivial since the matrix $M$ now relies on a group of weights $\exp(-E(n)/T)$ which need to be learned and updated. To reduce the high computational complexity of diagonalizing a large matrix,
we apply an architecture similar to GCN \cite{gcn} by circumventing the diagonalization of the matrix $M$, and directly multiply it to the left hand side of the input and accompany it by multiplying another weight matrix $W$ at the right hand side. The convolutional layer would then be reduced to a simple form
\begin{equation} \label{eq:conv}
X^{(h+1)}=M^{(h)}X^{(h)}W^{(h)},
\end{equation}
where $h$ refers to the layer number. Applying $M$ to the input $X$ is essentially a weighted average among neighbors of a given node. Here we call the graph convolution induced by MET the \emph{MET convolution}.
The model \eqref{eq:conv} can be simplified further. Instead of learning the Boltzmann's factors which enter $k(n;i)$ through \eqref{eq:Propagator}, one may treat $k(n;i)$ as a constant $k(n)$ that is independent of nodes and learn it directly. This simplification circumvents normalizing the summation by $Z$ and eases the training process. The convolutional layer is then simplified as
\begin{equation} \label{eq:conv2}
X^{(h+1)}=\sum_{n=0}^{L}k^{(h)}(n)D_n^{-1}A^nX^{(h)}W^{(h)},
\end{equation}
where $D_n$ is the degree matrix for $A^n$. We call the graph convolutional networks in \eqref{eq:conv} and \eqref{eq:conv2} PAN as the path integral based MET convolution is used.

\begin{figure}[ht]
\begin{center}
\centering
\begin{minipage}{\columnwidth}
\begin{minipage}{0.32\columnwidth}
	\includegraphics[width=1\columnwidth]{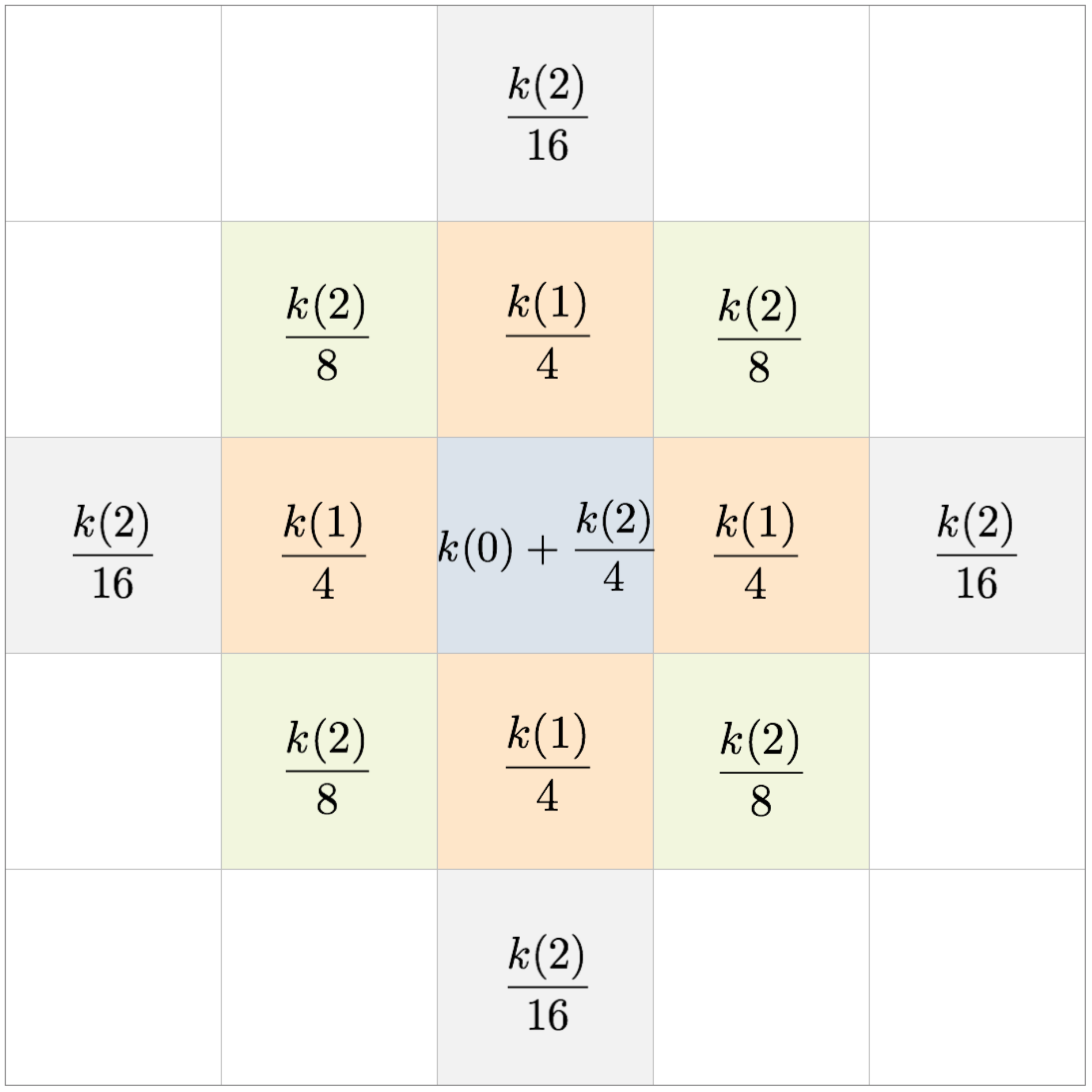}\\
	\centering
	{\small (a) Maximal entropy}
\end{minipage}
\begin{minipage}{0.32\columnwidth}
	\includegraphics[width=1\columnwidth]{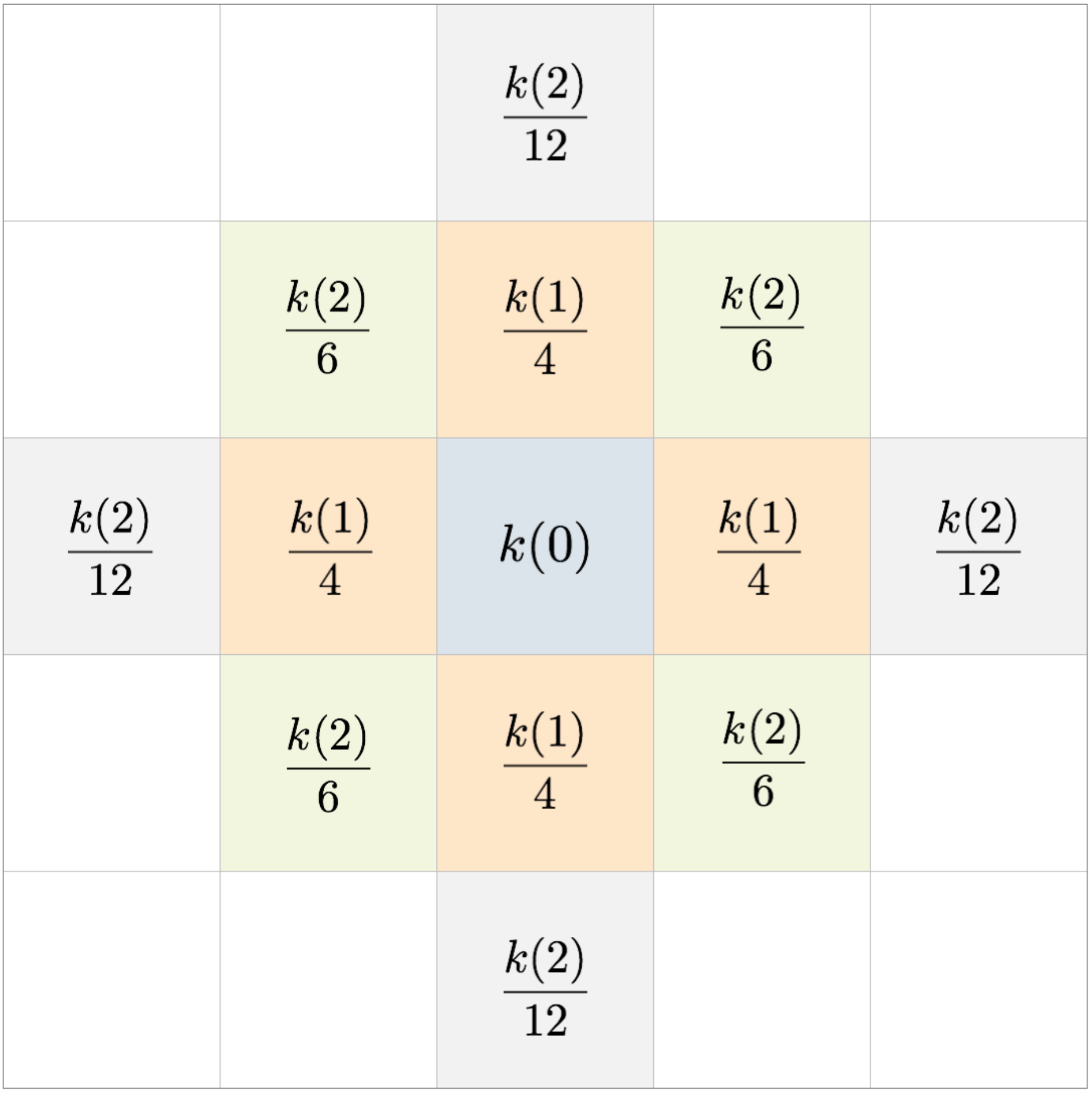}\\
\centering	
	{\small (b) Shortest path}
	\end{minipage}
\begin{minipage}{0.32\columnwidth}
	\includegraphics[width=1\columnwidth]{Cov_short}\\
	\centering
	{\small (c) Self-avoiding}
\end{minipage}
\end{minipage}
\caption{Mapping model \eqref{eq:conv2} ($L=2$) to a traditional convolution filter for different types of paths. (a) Maximal entropy paths. (b) Shortest paths. (c) Self-avoiding paths. (b) and (c) happen to be the same for $L=2$.}
\label{fig: 3times3}
\end{center}
\end{figure}

Note that this simplified model can only be equal to \eqref{eq:conv} when the graph is regular. Interestingly, if one interprets an image as a regular graph, where pixels with shared edges are considered connected, an analog can be drawn between the model \eqref{eq:conv2} and traditional CNNs. For traditional CNNs, elements of a convolution filter can be associated with the Cartesian coordinates of the neighbors of a given node. For irregular graphs, this coordinate system is not available. However, one can still associate the filter elements (i.e., $k(n)$) with the neighbors by the scalar quantity ``distance''.
As an example, we explicitly map model \eqref{eq:conv2} to a traditional convolution filter in $\mathbb{R}^2$ in Figure~\ref{fig: 3times3}. We present the mapping for three different types of paths, although in this paper we only focus on the maximal entropy one due to its simple form of $g(i,j;n)$.
\section{Related work}
The maximal entropy random walk has already shown excellent performance on link prediction \cite{li2011link} or community detection \cite{ochab2013maximal} tasks.
Essentially, most graph Laplacian based models \cite{N-GCN2018,DCNN_2016,BrZaSzLe2013,ChebNet,gcn,MoNet_2016,Graph-CNN_2017,GWNN} can be adapted to our framework by replacing the graph Laplacian with the MET matrix $M$. This represents a change from the GRW model \cite{DeepWalk_2014} or its modified versions \cite{tang2015line,node2vec_2016} which sample random walks by local information or the similarity of nodes, to a global information based parameter-free random walk.
Many popular models can be related to or viewed as certain explicit realizations of our framework. Besides the direct link between our model and traditional CNNs mentioned above,  the MET matrix can also be interpreted as an operator that acts on the graph input, which works as a kernel that allocates appropriate weights among the neighbors of a given node. This mechanism is similar to the attention mechanism \cite{gat}, while we restrict the functional form of $M$ based on physical intuitions. Specifically, we suppose the operator can be expanded as a series $\sum_{n=0}^{r} f_n(A,X)A^n$, it then performs the attention mechanism but preserves a compact form. Although we keep the number of features by applying $M$, one can easily concatenate the aggregated information of neighbors like GraphSAGE \cite{hamilton2017} or GAT \cite{gat}. The optimal order $r$ of the series depends on the intrinsic properties of the graph, which is represented by temperature $T$. Incorporating more terms is analogous to having more particles excited to higher energy level at higher temperature. For instance, in the \emph{low-temperature limit}, $M=I$, the model is reduced to the MLP model. In the \emph{high-temperature limit}, all factors $\exp(-E(n)/T)$ become effectively one, and the summation is dominated by the term with the largest power. This can be seen by noticing 
\begin{equation}
A^n=\sum_{i=1}^{N}\lambda_i^n \psi_i\psi_i^T,
\end{equation}
where $\lambda_1,\dots,\lambda_N$ is sorted in a descending order. By the Perron-Frobenius theorem we may only keep the leading order term with the unique largest eigenvalue $\lambda_1$ when $n\rightarrow \infty$. We then reach a prototype of the high temperature model $X^{(h+1)}=(I+\psi_1\psi_1^T)X^{(h)}W^{(h)}$. Empirically, a moderate order of one to three seems to perform better than both extremes, which well reflects the intrinsic dynamics of the graph. In particular, by choosing $L=1$ and $E(0)=E(1)=0$, model \eqref{eq:conv2} is essentially the GCN model \cite{gcn}. The trick of adding self-loops is automatically realized in higher powers of $A$. By replacing $A$ in \eqref{eq:conv2} or \eqref{eq:conv} with $D^{-1}A$ or $D^{-\frac{1}{2}}AD^{-\frac{1}{2}}$, we can easily transform our model to a multi-step GRW version, which is indeed the format of LanczosNet \cite{LNet}. Moreover, Lanczos algorithm may be directly applied to the MET matrix once it is symmetrically normalized.


\section{Experiments}
\subsection{Datasets and Baselines}
We test our method on three public available citation graph datasets: Citeseer, Cora and Pubmed, with the fixed data
splits performed in \cite{Planetoid_2016,gcn}, in comparison with some existing methods including node2vec \cite{node2vec_2016}, Planetoid \cite{Planetoid_2016}, skip-gram based graph embeddings (DeepWalk) \cite{DeepWalk_2014}, ChebNet \cite{ChebNet}, GCN \cite{gcn} (together with some baselines as compared in their work), attention-based models such as AGNN \cite{Attention_GCN} and GAT \cite{gat}, deep graph infomax (DGI) \cite{DGI}, Bootstrap \cite{Bootstrap_2017}, multi-scale deep graph convolutional networks AdaLNet \cite{LNet}, simplified GCN model (SGC) \cite{SGC}, and graph wavelet neural network (GWNN) \cite{GWNN}.

\subsection{Results and Discussion}
In Table~\ref{tab1:results}, we present partial results for performance comparison. See Supplementary Material for a full comparison list and the detailed experimental setup.
The records are averaged classification accuracies (in percentage \%) of $10$ independent trials.
It can be observed that our methods for semi-supervised node classification outperforms most of the existing models, especially for Pubmed. For Cora, the obtained accuracy is slightly less than DGI, AGNN, GAT and GWNN, but higher than all the other models. For Citeseer, our result outperforms most of the models, apart from DGI, SGC, GWNN, GAT. For Pubmed, our model has a better performance than all other models but AGNN. Overall, the performance of our model can be placed in the top five among all models.
\begin{table}[h]
\caption{Performance comparison of PAN and some previously published models for Cora, Citeseer and Pubmed.}\label{tab1:results}
\begin{center}
{\tabcolsep=0pt\def\arraystretch{0.9}
\begin{tabularx}{230pt}{l *4{>{\Centering}X}}
\toprule
{\bf Method} & {\bf Cora} & {\bf Citeseer}  & {\bf Pubmed} \\ \midrule
node2vec & 74.9 &54.7 & 75.3 \\
Planetoid  & 75.7  & 64.7 & 77.2  \\
DeepWalk  & 67.2 &43.2 & 65.3 \\
ChebNet & 81.2  &69.8& 74.4\\
GCN  & 81.5  & 70.3 & 79.0  \\
AGNN & 83.1 &71.1& 79.9\\
GAT & 83.0  & 72.5 & 79.0  \\
Bootstrap  & 78.4 & 53.6& 78.8\\
DGI & 82.3 & 71.8& 76.8\\
AdaLNet & 80.4 & 68.7& 78.1\\
SGC & 81.0 & 71.9& 78.9\\
GWNN &82.8 &71.7 &79.1 \\
\midrule
\textbf{PAN} ($L=2$)&\textbf{82.0} &\textbf{71.2} &\textbf{79.2}  \\
\bottomrule
\end{tabularx}}
\end{center}
\vskip -0.1in
\end{table}
\begin{figure}[htbp!]
\vskip -0.08in
\begin{center}
\begin{minipage}{\columnwidth}
\hskip 0.1in
\begin{minipage}{0.92\columnwidth}
	\includegraphics[width=\columnwidth]{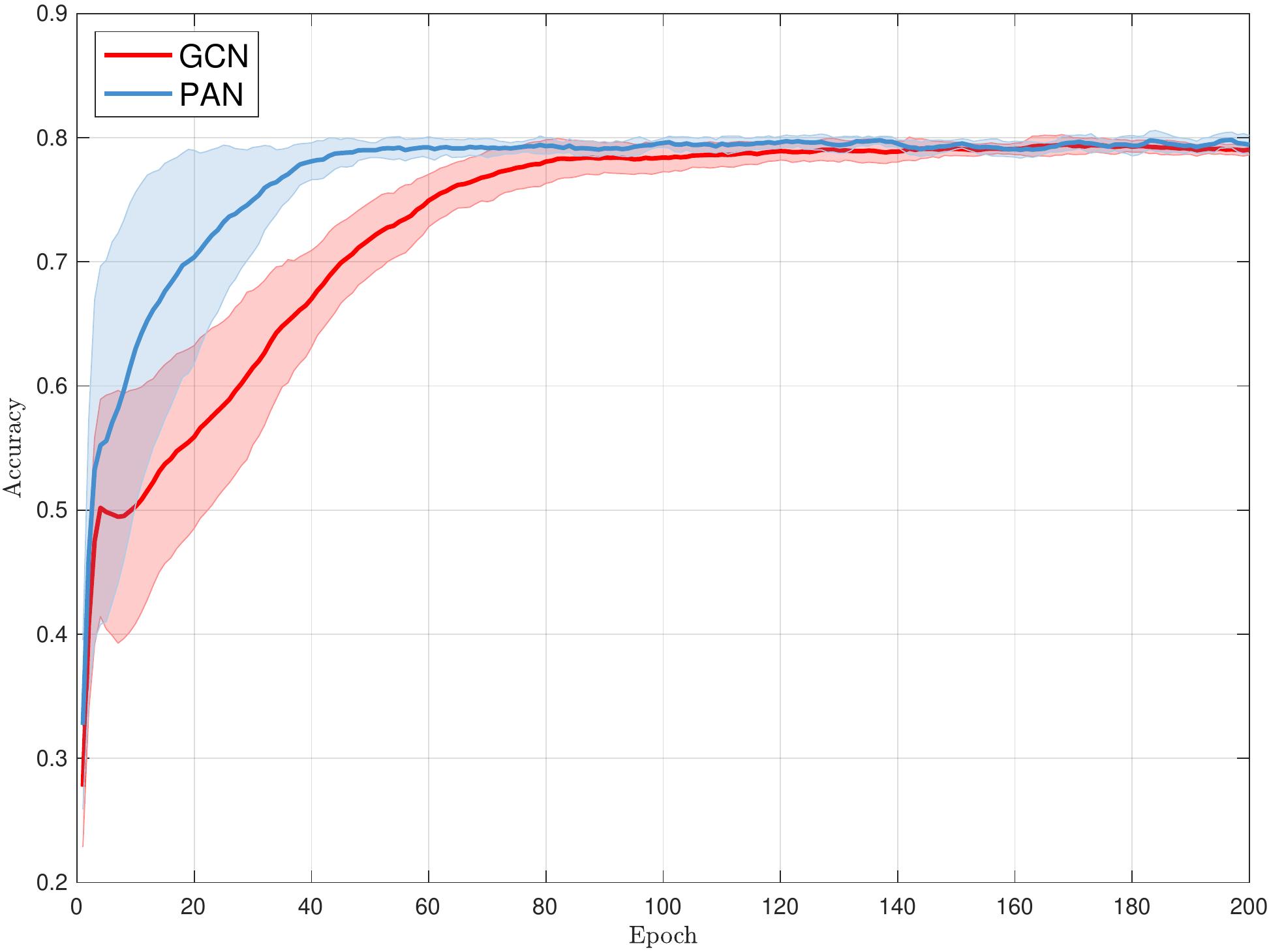}
\end{minipage}
	\hskip -0.5\columnwidth
\begin{minipage}{0.47\columnwidth}
\vskip 1.7cm
	\includegraphics[width=\columnwidth]{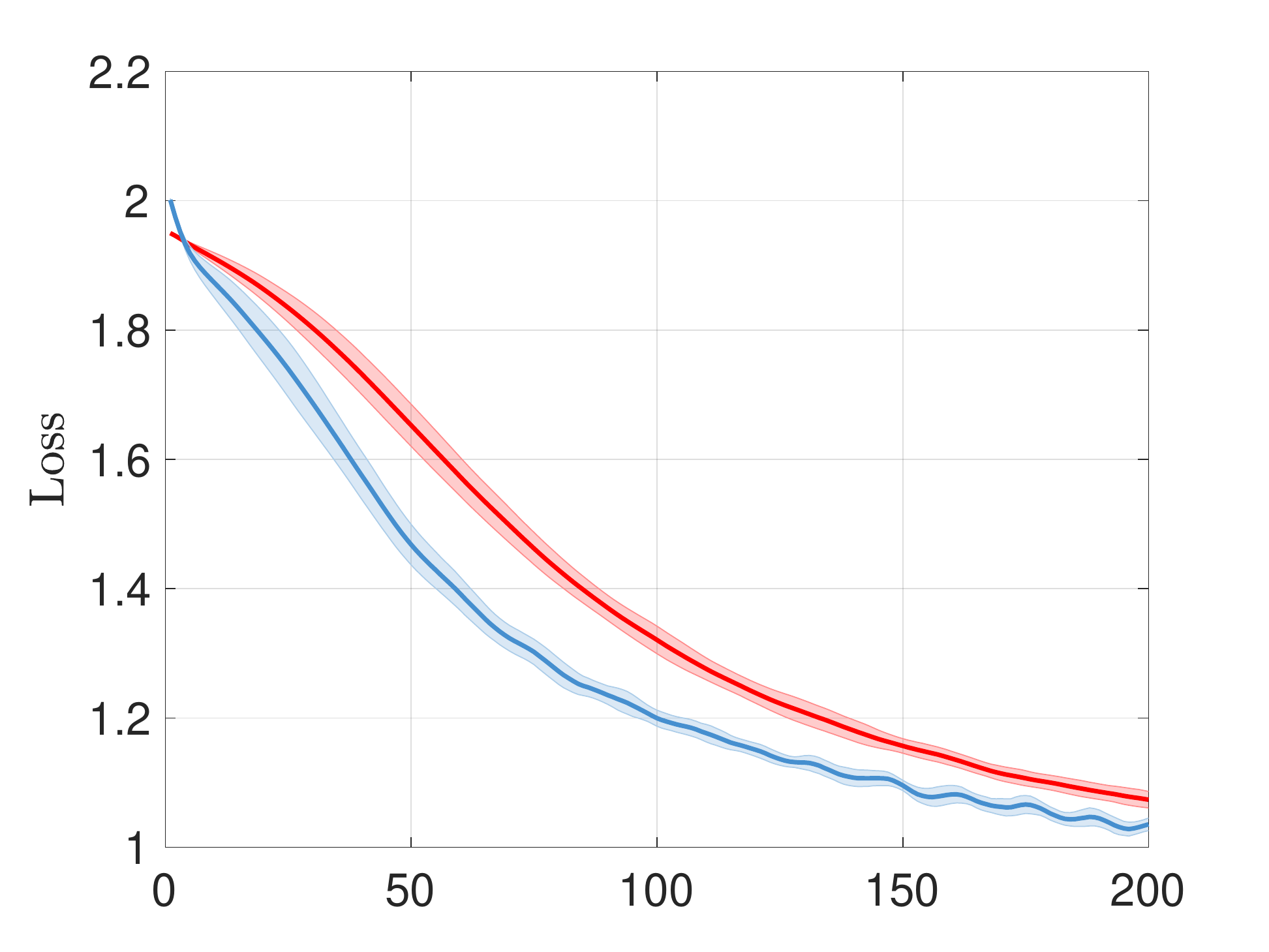}
\end{minipage}
\end{minipage}
\vskip -0.1in
\caption{Main figure: Mean and standard deviation of validation accuracies of PAN and GCN on Cora. Figure in lower right corner: Validation loss function of PAN and GCN.}
\label{fig:meanstd.PAN}
\end{center}
\vskip -0.2in
\end{figure}

We plot the trend of validation loss and accuracy (with mean and standard deviations) for both PAN and GCN on Cora in Figure \ref{fig:meanstd.PAN}. It illustrates that PAN converges faster and reaches higher accuracy compared to GCN, and finally obtains a better test performance as shown in Table \ref{tab1:results}. For model \eqref{eq:conv2}, a grid search of $k(n)$ may be more reliable than the gradient descent if $L$ is not large. The generalization ability may be improved by further constraining the number of parameters. For example, one may assume a certain form of $E(n)$, such as $E(n) \propto n^{\alpha}$ with $\alpha \geq 1$, and only train the temperature.

In general, whether maximal entropy random walk or GRW based model performs better depends on the nature of the graph. GRW may underestimate the contribution of the ``influencer" in a small-world network \cite{NeBaWa2011}, due to the fact that information sent from an ``influencer" is diluted as a result of the large degree it has \cite{Kampffmeyer_etal2019}. But in the maximal entropy random walk model, information transmitted in a path from the sender to the receiver is not affected by the degree of the sender.

\section{Conclusions}
We present a new graph convolution framework based on the path integral idea, which realizes the attention-like mechanism while preserves the simple form similar to GCN. Although we focus on maximal entropy random walk, our framework can easily accommodate other types of walks including many previous models. Preliminary results on node classification tasks show that our method achieves state-of-the-art performance very efficiently. Many extensions of the present work can be expected, including those for graph classification tasks.




\bibliography{references}
\bibliographystyle{icml2019}

\newpage
\appendix
\section{Experiments}
\paragraph{Full list of baselines.}
We test our method on three public available citation graph datasets: Citeseer, Cora and Pubmed, with the fixed data
splits performed in \cite{Planetoid_2016,gcn}. We compare our proposed model with 24 existing methods: MoNet \cite{MoNet_2016}, node2vec \cite{node2vec_2016}, Diffusion-CNN (DCNN) \cite{DCNN_2016}, GCN model \cite{gcn} and some baselines as compared in their work: MLP, ChebNet \cite{ChebNet}, label propagation (LP) \cite{LP_2003}, manifold regularization (ManiReg) \cite{ManiReg_2006}, semi-supervised embedding (SemiEmb) \cite{SemiEmb_2012}, skip-gram based graph embeddings (DeepWalk) \cite{DeepWalk_2014}, the iterative classification algorithm (ICA) \cite{ICA_30} and Planetoid \cite{Planetoid_2016}. To make a comprehensive comparison, we also include the recent proposed Graph-CNN \cite{Graph-CNN_2017}, DynamicFilter \cite{DynamicFilter_2017}, FastGCN \cite{Fast-GCN}, graph linear network (GLN) \cite{Attention_GCN}, attention-based models, such as AGNN \cite{Attention_GCN} and GAT \cite{gat}, deep graph infomax (DGI) \cite{DGI}, Bootstrap \cite{Bootstrap_2017}, multi-scale deep graph convolutional networks such as LNet and AdaLNet \cite{LNet}, simplified GCN model (SGC) \cite{SGC}, and graph wavelet neural network (GWNN) \cite{GWNN}.

\paragraph{Experimental settings.}
In our experiments, we apply the same model architecture as used in \cite{gcn}, i.e., a two-layer model with 16 hidden neurons in the first hidden layer.
Our models for all these three datasets are trained by Adam SGD optimizer \cite{Adam} with an initial learning rate $0.01$, where Glorot strategy \cite{Glorot2010} is used for weights initialization. The maximum number of epochs for Cora and Citeseer is $200$, while for Pubmed is $100$. Dropout rate is set as $0.5$ for Cora and Citeseer while $0.4$ for Pubmed. Weight decay is set as 5e-3 for Cora, 1e-2 for Citeseer, 3e-3 for Pubmed. We use the same early stopping strategy on validation loss \cite{gcn} with a patience of $50$ epochs for Cora and Citeseer, and $15$ epochs for Pubmed. Results of PAN in Table \ref{tab1:results} and \ref{tab2:results}  are obtained based on the above parameter setting.
\begin{table*}[t]
\caption{Summary of results in terms of classification accuracies in percentage (\%), for Cora, Citeseer and Pubmed, with a fixed (public) split of data from \cite{gcn}. `-' stands for the missing values when the existing work does not test the associated dataset.}\label{tab2:results}
\begin{center}
{\tabcolsep=0pt\def\arraystretch{1.2}
\begin{tabularx}{460pt}{l *8{>{\Centering}X}}
\toprule
{\bf Method} & {\bf Cora} & {\bf Citeseer}  & {\bf Pubmed} &\hspace{4mm}{\bf Method} & {\bf Cora} & {\bf Citeseer}  & {\bf Pubmed}\\
\midrule
LP  & 68.0  &45.3& 45.3      &Graph-CNN  & 76.3  &- &- \\
ICA  & 75.1  & 69.1& 73.9    & DynamicFilter  & 81.6  &- & 79.0 \\
ManiReg & 59.5  &60.1 & 70.7  &FastGCN  &81.8  &- &77.6\\
SemiEmb  & 59.0  &59.6 & 71.7 &GLN  & 81.2 & 70.9& 78.9\\
DeepWalk  & 67.2 &43.2 & 65.3 &AGNN & 83.1 &71.1& 79.9\\
ChebNet & 81.2  &69.8& 74.4   &GAT & 83.0  & 72.5 & 79.0\\
DCNN & 76.8 &- & 73.0           & Bootstrap  & 78.4 & 53.6& 78.8\\
node2vec & 74.9 &54.7 & 75.3  &DGI & 82.3 & 71.8& 76.8\\
Planetoid  & 75.7  & 64.7 & 77.2  & LNet & 79.5 & 66.2& 78.3\\
MoNet  & 81.7 &- & 78.8            & AdaLNet & 80.4 & 68.7& 78.1\\
MLP         &55.1   & 46.5 & 71.4  & SGC & 81.0 & 71.9& 78.9\\
GCN  & 81.5  & 70.3 & 79.0        & GWNN &82.8 &71.7 &79.1 \\
\midrule
\textbf{PAN}& \textbf{82.0} &\textbf{71.2} & \textbf{79.2} & & & &  \\
\bottomrule
\end{tabularx}}
\end{center}
\end{table*}

\paragraph{Variations of PAN.}
We test the performance of some variations of PAN. Depending on the normalization method for adjacency matrix $A$ and $D_n$ which
 is the degree matrix for $A^{n}$, we propose totally seven different versions of PAN in (9). In the following, $\tilde{D}_n$ is the degree matrix for $\tilde{A}^{n}$ with $\tilde{A}=A+I$, $\hat{A}=\tilde{D}^{-1/2}\tilde{A}\tilde{D}^{-1/2}$, and $\tilde{D}$ is the degree matrix for $\tilde{A}$.
\paragraph{Method 1}
\begin{equation*}
X^{(h+1)}=Z^{-1}\sum_{n=0}^{L}e^{-\frac{E(n)}{T}}A^nX^{(h)}W^{(h)}
\end{equation*}
\paragraph{Method 2}
 \begin{equation*}
X^{(h+1)}=Z^{-1/2}\sum_{n=0}^{L}e^{-\frac{E(n)}{T}}A^nZ^{-1/2}X^{(h)}W^{(h)}
\end{equation*}
\paragraph{Method 3}
\begin{equation*}
X^{(h+1)}=\sum_{n=0}^{L}k^{(h)}(n)D_n^{-1}A^nX^{(h)}W^{(h)}
\end{equation*}
\paragraph{Method 4}
\begin{equation*}
X^{(h+1)}=\sum_{n=0}^{L}k^{(h)}(n)\tilde{D}_n^{-1}\tilde{A}^nX^{(h)}W^{(h)}
\end{equation*}
 \paragraph{Method 5}
 \begin{equation*}
X^{(h+1)}=\sum_{n=0}^{L}k^{(h)}(n)\hat{A}^nX^{(h)}W^{(h)}
\end{equation*}
\paragraph{Method 6}
\begin{equation*}
X^{(h+1)}=\sum_{n=0}^{L}k^{(h)}(n)D_n^{-1/2}A^nD_n^{-1/2}X^{(h)}W^{(h)}
\end{equation*}
\paragraph{Method 7}
\begin{equation*}
X^{(h+1)}=Z^{-1}\sum_{n=0}^{L}e^{-\frac{E(n)}{T}}\hat{A}^nX^{(h)}W^{(h)}
\end{equation*}
The results of Methods 1--7 are shown in Table~\ref{tab:variationsofPan}, where backpropagation (BP) algorithm is used for training the weights $k^{(h)}(n)$ while in the bottom line we use grid search to seek for the optimal $k^{(h)}(n)$. Based on our empirical study, the highest accuracies for all these datasets are achieved at $k^{(h)}(0)=0$, $k^{(h)}(1)=k^{(h)}(2)$, as shown in the last row of Table ~\ref{tab:variationsofPan}. Method 2 also appears to be a strong candidate. More theoretical analysis on why these models are favourable is expected in our future work. 

\begin{table}[h]
\caption{Performance comparison for Method 1-7 on Cora, Citeseer and Pubmed with $L=2$.}\label{tab:variationsofPan}
\begin{center}
{\tabcolsep=0pt\def\arraystretch{1.2}
\begin{tabularx}{230pt}{l *4{>{\Centering}X}}
\toprule
{\bf Method} & {\bf Cora} & {\bf Citeseer}  & {\bf Pubmed} \\ \midrule
Method 1 & 80.7 & 69.2 &  77.7\\
Method 2  & 81.3 &70.1 & 77.7 \\
Method 3   &80.8  &69.2 & 78.8 \\
Method 4 & 80.9 &68.9 &78.8  \\
Method 5 & 79.8 &66.6 & 75.3 \\
Method 6 & 80.1 & 68.5& 75.8 \\
Method 7 & 80.8 & 69.6& 78.4 \\
\midrule
Method 5 (grid search) &\textbf{82.0} &\textbf{71.2} &\textbf{79.2}  \\
\bottomrule
\end{tabularx}}
\end{center}
\vskip -0.1in
\end{table}


\end{document}